\documentclass[runningheads]{llncs}
\usepackage[T1]{fontenc}
\usepackage{multirow}
\usepackage{caption} 
\usepackage{booktabs}
\usepackage{graphicx}   % For resizing tables
\usepackage{floatrow}   % For placing tables side by side
\usepackage{booktabs}
\usepackage{floatrow}
\newfloatcommand{capbtabbox}{table}[][\FBwidth]
\usepackage{multirow}   % For multi-row cells
\usepackage{rotating}   % For rotating text
\usepackage{graphicx,verbatim}
\usepackage{amsmath} 
\usepackage{pifont}
\newcommand{\xmark}{\ding{53}}%
\usepackage{amssymb}
\usepackage{float} 
\usepackage[table]{xcolor}
\usepackage{amssymb}  % For \checkmark
\usepackage{hyperref}
\usepackage{caption}
\usepackage[misc]{ifsym}

\begin{document}
\title{Endo-FASt3r: Endoscopic Foundation model Adaptation for Structure from motion}
%
\begin{comment}  %% Removed for anonymized MICCAI 2025 submission
\author{First Author\inst{1}\orcidID{0000-1111-2222-3333}\and
Second Author\inst{2,3}\orcidID{1111-2222-3333-4444} \and
Third Author\inst{3}\orcidID{2222--3333-4444-5555}}
%
\authorrunning{F. Author et al.}
\institute{Princeton University, Princeton NJ 08544, USA \and
Springer Heidelberg, Tiergartenstr. 17, 69121 Heidelberg, Germany
\email{lncs@springer.com}\\
\url{http://www.springer.com/gp/computer-science/lncs} \and
ABC Institute, Rupert-Karls-University Heidelberg, Heidelberg, Germany\\
\email{\{abc,lncs\}@uni-heidelberg.de}}

\end{comment}

% \author{Mona Sheikh Zeinoddin, Mobarakol Islam, Zafer Tandogdu, Greg Shaw, Mathew J. Clarkson, Evangelos Mazomenos, Danail Stoyanov}  %% Added for anonymized MICCAI 2025 submission
% \authorrunning{Mona Sheikh Zeinoddin et al.}
% \institute{University College London \\
%     \email{mona.zeinoddin.22@ucl.ac.uk}}

\author{
    Mona Sheikh Zeinoddin\inst{1,2}\textsuperscript{(\Letter)} \and
    Mobarak I. Hoque\inst{1,4} \and
    Zafer Tandogdu\inst{3,6} \and
    Greg L. Shaw\inst{3} \and
    Matthew J. Clarkson\inst{1,4} \and
    Evangelos B. Mazomenos\inst{1,4}\textsuperscript{(\Letter)} \and
    Danail Stoyanov\inst{1,5}
}
% index{Sheikh Zeinoddin, Mona}
% index{Hoque, Mobarak}
% index{Tandogdu, Zafer}
% index{Shaw, Greg}
% index{Clarkson, Matthew}
% index{Mazomenos, Evangelos}
% index{Stoyanov, Danail}

\authorrunning{Mona Sheikh Zeinoddin et al.}

\institute{
    UCL Hawkes Institute, University College London, UK \\ 
    \email{\{mona.zeinoddin.22; e.mazomenos\}@ucl.ac.uk}\and
    Institute of Health Informatics, University College London, UK \and
    Dept of Urology, University College London Hospitals, UK \and
    Dept of Medical Physics \& Biomedical Engineering, University College London, UK \and
    Dept of Computer Science, University College London, UK \and
    Division of Surgery and Interventional Science, University College London, UK \\
}    
\maketitle
\begin{abstract}
Accurate depth and camera pose estimation is essential for achieving high-quality 3D visualisations in robotic-assisted surgery. Despite recent advancements in foundation model adaptation to monocular depth estimation of endoscopic scenes via self-supervised learning (SSL), no prior work has explored their use for pose estimation. These methods rely on low rank-based adaptation approaches, which constrain model updates to a low-rank space. We propose Endo-FASt3r, the first monocular SSL depth and pose estimation framework that uses foundation models for both tasks. We extend the Reloc3r relative pose estimation foundation model by designing Reloc3rX, introducing modifications necessary for convergence in SSL. We also present DoMoRA, a novel adaptation technique that enables higher-rank updates and faster convergence. Experiments on the SCARED dataset show that Endo-FASt3r achieves a substantial $10\%$ improvement in pose estimation and a $2\%$ improvement in depth estimation over prior work. Similar performance gains on the Hamlyn and StereoMIS datasets reinforce the generalisability of Endo-FASt3r across different datasets.  Our code is available at: \href{https://github.com/Mona-ShZeinoddin/Endo_FASt3r.git}{https://github.com/Mona-ShZeinoddin/Endo\_FASt3r.git}.

% Joint depth and camera pose estimation are fundamental in robotic-assisted surgery.
% This work advances the low-rank adaptation of foundation models to endoscopic scenes following self-supervised learning (SSL). We propose EndoFASt3r, a monocular SSL joint depth and pose estimation framework that, for the first time, leverages foundation models for both tasks. We extend the Reloc3r relative pose estimation foundation model by introducing Reloc3rX, a modification necessary for convergence in SSL. Furthermore, we introduce DoMoRA, a novel adaptation technique that enables updates in a higher-rank space while achieving faster convergence. Experiments on the SCARED dataset indicate that Endo-FASt3r achieves a substantial $10\%$ improvement in pose estimation, and a $2\%$ improvement in depth estimation over state-of-the-art methods. Further evaluation on the Hamlyn dataset confirms similar performance gains, reinforcing the generalisability of Endo-FASt3r across different surgical datasets.


\end{abstract}

\section{Introduction}
Accurate depth and camera pose estimation are prerequisites to gaining a deeper geometric understanding of robotic-assisted surgery (RAS) scenes~\cite{qian2019review}. However, the unique properties of RAS such as varying illumination, textureless areas, and frequent occlusions complicate these tasks~\cite{chen2018slam}. Although deep learning-based methods are a common approach~\cite{ali2022we}, their development is limited by the lack of large-scale annotated endoscopic datasets, driving the adoption of self-supervised learning (SSL). With the rise of foundation models~\cite{damv2,dinov2}, recent works have explored their potential in RAS~\cite{surgicaldino,dares}. However, directly applying them to RAS is suboptimal~\cite{dares}, inspiring the introduction of adaptation techniques. The potential of these techniques and the need for SSL have motivated the adaptation of foundation models to RAS using SSL. While many SSL monocular depth and pose estimation methods in RAS
have utilised foundation models in their depth module
~\cite{surgicaldino,dares}, none have explored their use in the pose module. Leveraging a foundation model in
the pose module can address key challenges, such as variations in camera
movement and occlusions, which often lead to suboptimal performance in
current CNN-based pose modules that focus mainly on local patterns~\cite{schmidt2024tracking,zhong2025review}. Recent works on parameter-efficient fine-tuning techniques (PEFT) in SSL monocular depth estimation of RAS scenes~\cite{dares,surgicaldino} have been limited by the use of low-rank updates driven by the low-rank adaptation technique (LoRA)~\cite{lora}. While LoRA’s approach of decomposing the weight update matrix into low-rank matrices is effective, it constrains adaptability, particularly in cases with significant domain shifts, such as RAS, where higher-rank updates are essential for optimal performance~\cite{mora}. We introduce Endo-FASt3r, a novel SSL monocular depth and pose
estimation framework that relies on foundation models for both tasks, making Endo-FASt3r the first to adapt a foundation model for pose estimation in RAS scenes. For the pose module, we design Reloc3rX, extending the Reloc3r~\cite{reloc3r} foundation model for robust convergence in SSL. We also present DoMoRA, a novel adaptation technique that advances recent PEFTs~\cite{mora,dora} by incorporating both low-rank and full-rank updates while benefiting from faster convergence. Our main contributions and findings include:

% We introduce Endo-FASt3r, a novel SSL monocular depth and relative pose estimation technique that relies on foundation models (dav2, Reloc3r) for both tasks. In Endo-FASt3r we are the first to adapt a foundation model for pose estimation in RAS scenes. For the pose module, we design Reloc3rX, modifying the Reloc3r~\cite{reloc3r} foundation model for robust convergence in SSL. Additionally, we present DoMoRA, a novel adaptation technique that extends recent PEFTs~\cite{mora,dora} by considering the necessity of both low-rank and full-rank updates while benefiting from faster convergence. Our main contributions and findings include:
\begin{enumerate}
\item We design Reloc3rX by extending~\cite{reloc3r}, which addresses the scale sensitivity issue of monocular SSL-based depth and pose estimation. Our work is the first to use a foundation model for pose estimation in RAS.  

\item We propose DoMoRA, a novel adaptation technique that enhances previous PEFTs by addressing the necessity of both full-rank and low-rank updates while improving convergence through weight decomposition.

\item Experimental results on three public datasets, SCARED, Hamlyn, and, StereoMIS show a substantial improvement ranging from $7\%-10\%$ in the absolute trajectory error metric for pose estimation and a $2\%$ improvement in the absolute relative error metric for depth estimation over state-of-the-art (SOTA) methods, highlighting the benefits of Endo-FASt3r.
% foundation models for both modules, in addition to our DoMoRA technique. 
\end{enumerate}

\section{Methods}\label{sec2}
\subsection{Preliminaries}
\noindent \textbf{Depth Anything (DA) V2:} 
\label{dav2}
Previously trained on 62M urban scene images, DA V2 employs a 12-layer DINO V2-based encoder~\cite{dinov2} and a decoder with convolutional head and neck modules~\cite{damv2}. This work aims to better adapt DA V2 for depth estimation in RAS through the proposed adaptation technique, DoMoRA. 
\newline
\newline
\noindent \textbf{Reloc3r:}
\label{reloc3r}
% \subsubsection{Reloc3r} 
To directly regress the relative pose between image pairs~\cite{reloc3r}, Reloc3r processes each image $I_i$ through 24 ViT encoder blocks, producing features $F_1$ and $F_2$ for each image. The decoder includes 12 ViT blocks, each integrating a cross-attention layer between its self-attention and feed-forward layers to enhance spatial alignment between $F_1$ and $F_2$. The two branches are fully symmetrical with shared weights. The decoder produces features $G_{1}$ and $G_{2}$, which are then fed into a pose regression head comprising 2 feed-forward layers to regress the 9D intermediate representation of rotation and 3D translation from $I_1$ to $I_2$ and $I_2$ to $I_1$:
\begin{equation}
\label{relocer-head}
\hat{R}_{I_1, I_2}^{(3 \times 3)}, {t}_{I_1, I_2}^{(3 \times 1)} = \text{Head}(G_{1}), \ \ \ 
\hat{R}_{I_2, I_1}^{(3 \times 3)} , {t}_{I_2, I_1}^{(3 \times 1)} = \text{Head}(G_{2}) 
\end{equation}

The 9D intermediate rotation representation is then orthogonalised using singular value decomposition into a $3 \times 3$ rotation matrix, denoted as function $Orthogonal$.

\begin{equation}
\label{ortho}
{R}_{I_1, I_2}^{(3 \times 3)} = \text{Orthogonal}(\hat{R}_{I_1, I_2}^{(3 \times 3)}), \ \ \ 
{R}_{I_2, I_1}^{(3 \times 3)} = \text{Orthogonal}(\hat{R}_{I_2, I_1}^{(3 \times 3)}) 
\end{equation}

Reloc3r was trained on 6M images in a supervised manner via the angular distance between the predicted and ground truth poses as the supervision signal. However, we have empirically verified that directly using Reloc3r in an SSL framework for RAS scenes causes divergence. Our work focuses on redesigning Reloc3r's pose head to account for scale sensitivity, making Endo-FASt3r the first framework to adapt a foundation model for pose estimation in RAS scenes. 
\newline
\newline
\noindent \textbf{MoRA:}
Improving upon LoRA-based techniques by using a square matrix as opposed to low-rank matrices, MoRA enhances the capacity of the weight update matrix ~\cite{mora}. Given the pre-trained weight $W_0 \in \mathbb{R}^{d \times k}$, MoRA allocates a square matrix $M \in \mathbb{R}^{r \times r}$ with rank r and defines the new layer with MoRA as:
\begin{equation}
\label{eq:mora}
\begin{split}
  %\hat{x} &= f_{\text{comp}}(x)\\
  %h &= W_0x + f_{\text{decomp}}(M\hat{x})
  h(x) &= W_0 x + f_{\text{decomp}}\Big(M  f_{\text{comp}}\left(x\right)\Big)
\end{split}
\end{equation}
% In which $f_{\text{comp}}: \mathbb{R}^k \rightarrow \mathbb{R}^{\hat{r}}$ is the function that compresses the input dimension from $k$ to $\hat{r}$, and $f_{\text{decomp}}: \mathbb{R}^{\hat{r}} \rightarrow \mathbb{R}^{d}$ is the function that decompresses the output dimension from $\hat{r}$ to $d$. These two functions are non-parameterized operators and are executed in linear time w.r.t to the dimension. 

In which $f_{\text{comp}}: \mathbb{R}^k \rightarrow \mathbb{R}^{r}$ and $f_{\text{decomp}}: \mathbb{R}^{r} \rightarrow \mathbb{R}^{d}$ are non-parameterized functions that compress the input dimension from $k$ to $r$  and decompress the output dimension from $r$ to $d$ respectively.
\newline
\newline
\noindent \textbf{DoRA:} 
Proposed to improve LoRA by drawing influence from Weight Normalization~\cite{weight}, DoRA decomposes the weight matrix into its magnitude and direction to accelerate optimisation~\cite{dora}. The weight decomposition of $W \in \mathbb{R}^{d \times k}$ can be written as $W = m\frac{V}{||V||_{c}}$ where $m \in \mathbb{R}^{1 \times k}$ is the magnitude vector, $V \in \mathbb{R}^{d \times k}$ is the directional matrix, with $||\cdot||_{c}$ being the column-wise vector norm of a matrix. At initialisation, given the pre-trained weight $W_0$, we set $m = ||W_0||_{c}$ and $V = W_0$. LoRA with rank $r$ is then applied only to the directional component while the magnitude component is kept trainable. The updated weight $W'$ is:
\begin{equation}
W'= \underline{m}\frac{V+\Delta V}{||V+\Delta V||_{c}} = \underline{m}\frac{W_0+\underline{BA}}{||W_0+\underline{BA}||_{c}}
\label{eq:DoRA}
\end{equation}
where $\Delta V$ is the incremental directional update learned by multiplying two low-rank matrices $B$ and $A$, with the underlined parameters being trainable. The new layer with DoRA adaptation $h(x)$ can be defined as $h(x) = W' x$.
\newline
\newline
\noindent \textbf{Self-Supervised Depth and Pose Estimation:}
\label{ssl}
The reprojection loss-based approach~\cite{monodepth2} is a widely used SSL method in monocular depth and pose estimation, especially in RAS~\cite{afsfm}. It comprises a depth module, which estimates the depth map of the source frame $I_{s}$, and a pose module, which predicts the camera movement between the source frame $I_{s}$ and target frame $I_{t}$. Given the depth prediction $D_{s}$ for $I_{s}$, and $R$ and $t$, the rotation and translation matrices describing the camera movement, the synthetic target frame can be derived using the reprojection function $\pi$ as $I_{s\to t} = \pi \left (  D_{s},K,R,t,I_{s}\right )$, in which K is the camera intrinsics. A multi-scale SSIM ($\mathrm{MS\_SSIM}$)-based~\cite{multiscale} reprojection loss proposed in~\cite{dares} is used to measure the dissimilarity between the original and synthetic image:
\begin{equation}
L_{\text{reproj}} = \alpha \Big( 1 - \mathrm{MS\_SSIM}
(I_{\text{t}}, I_{s\to t}) \Big) + \beta \left|I_{\text{t}} - I_{s\to t} \right|
\end{equation}

To address the varying illumination conditions and occlusions in RAS scenes, we have followed the approach in~\cite{afsfm} which introduces the Tihkonov regulariser loss terms. The total loss $L_{\text{ssl}}$ can be written as $L_{\text{ssl}} = L_{\text{reproj}} + L_{\text{tihkonov}}$.
\begin{figure}[t]
\centering
\includegraphics[width=0.90\textwidth]{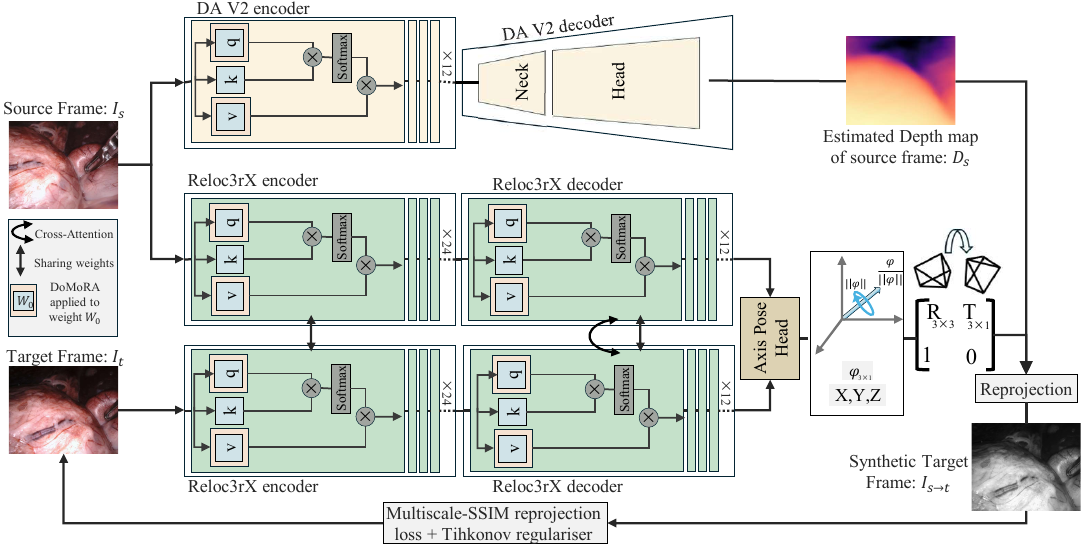}
\caption{EndoFAst3r SSL monocular depth and pose estimation pipeline based on the two foundation models, DA V2 and Reloc3rX.}
\label{main}
\end{figure}
\subsection{Proposed Framework: Endo-FASt3r}
As presented in Fig.~\ref{main}, Endo-FASt3r structures an SSL framework for monocular endoscopic depth and pose estimation by incorporating DA V2 in the depth module and Reloc3rX in the pose module. We propose a novel adaptation method, DoMoRA, to better adapt both DA V2 and Reloc3rX to RAS scenes. 
\newline
\newline
\noindent \textbf{DoMoRA:}
To benefit from the faster convergence of DoRA and the full-rank update space of MoRA, we introduce DoMoRA. Following Eqs.~\ref{eq:DoRA},~\ref{eq:mora} and given the pre-trained weight $W_0^{d \times k}$, rank $r$, magnitude $m$, DoRA low-rank matrices $B ^{d \times r}$,$A ^{r \times k}$ and MoRA square matrix $M^{r \times r}$, the new layer with DoMoRA is:

\begin{equation}
h(x) = (\underline{m}\frac{W_0+\underline{BA}}{||W_0+\underline{BA}||_{c}}) x + f_{\text{decomp}}\left(M  f_{\text{comp}}\Big(x\right)\Big)
\label{eq:DoMoRA}
\end{equation}

\begin{figure}
\centering
\includegraphics[width=0.45\textwidth]{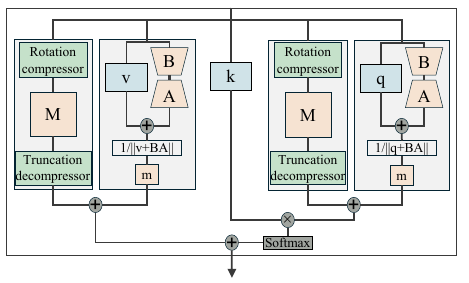}
\caption{Illustration of the DoMoRA Transformer block, in which matrices $M,m,A,B$ are trainable while pre-trained $q,k,v$ matrices are frozen. In addition, DA V2 decoder neck, head and Reloc3rX pose head are also trainable. } 
\label{domora_fig}
\end{figure}

In which $f_{\text{comp}}$  and $f_{\text{decomp}}$ follow the rotation and truncation methods\cite{mora}. Endo-FASt3r incorporates DoMoRA in the query and value matrices of all transformer blocks (Fig.~\ref{domora_fig}). Inspired by the hierarchical feature learning nature of neural networks, we adopt the vector rank approach presented in~\cite{dares}.
\newline
\noindent \textbf{Reloc3rX:}
We design Reloc3rX, an extension of Reloc3r tailored for RAS. Unlike supervised methods, SSL-based monocular depth and pose estimation suffers from scale ambiguity~\cite{scale}. This is caused by the scale-sensitive Tihkonov regulariser~\cite{afsfm} loss term, which leads to training instability and even divergence. We empirically verified that directly applying Reloc3r, pre-trained in a supervised manner on urban scenes, results in divergence during the early stages of training in RAS, due to scale misalignment. To address this, the pose regression head in Eq.~\ref{relocer-head} was modified to predict rotation in the 3D axis angle space instead of the 9D rotation matrix space, eliminating the need for orthogonalisation in Eq.~\ref{ortho}. To ensure alignment with RAS scale, particularly at the early stages of training, the predicted axis-angle and translation vectors are scaled by a factor of $0.001$. After scaling, the resulting axis angle ${\phi}_{I_1, I_2}^{(3 \times 1)}$is converted to a rotation matrix via the Rodrigues rotation formula~\cite{murray2017mathematical}.

\begin{equation}
{\phi}_{I_1, I_2}^{(3 \times 1)}, {t}_{I_1, I_2}^{(3 \times 1)} = \text{Head}(G_{1}^{(T \times d)}) 
% \hat{\phi}_{I_2, I_1}^{(3 \times 1)} , {t}_{I_2, I_1}^{(3 \times 1)} = \text{Head}(G_{2}^{(T \times d)}) 
\end{equation} 

Unlike Reloc3r, Reloc3rX computes the pose only from the source image to the target image, as the reverse is unnecessary for our SSL framework.
\newline
\newline
\noindent \textbf{Training and Evaluation framework:}
In the training phase, given the source frame $I_{s}$ and the the target frame $I_{t}$, the depth module operates on the source frame $I_{s}$ and produces its corresponding depth map. Meanwhile, the pose module will operate on both frames to estimate their relative camera pose. The synthetic target frame is then generated via reprojection. The Tihkonov regulariser and reprojection loss are used for optimisation. In the evaluation phase, each module operates independently to produce the desired depth map and camera pose. 

\section{Implementation details}
\noindent \textbf{Training:}
Endo-FASt3r utilises the Adam optimiser with a learning rate of $1e-4$ decaying every 10 steps. Training was performed on an A100 GPU for 10 epochs with a batch size of 4 for $20$ hours.
\newline
\newline
\noindent \textbf{Datasets and Evaluation Protocol:}
Endo-FASt3r has been trained and evaluated on the SCARED dataset ~\cite{scared}, captured with a da Vinci Xi from porcine cadavers using structured light for depth collection. Following the split in~\cite{afsfm}, 15,351 frames were used for training, 1,705 for validation, and 551 for depth evaluation. Meanwhile pose evaluation utilised two trajectories of length 410 and 833 frames defined in~\cite{afsfm}. To examine generalisability, we used the Hamlyn dataset following~\cite{endo-depth&motion} for the depth module. For the pose module, we extracted frames 9,780–10,980 from sequence P3 of the StereoMIS dataset~\cite{hayoz2023learning}, which exhibit significant tissue deformation and camera motion in an \textit{in vivo} setup. StereoMIS ground-truth poses were derived from the kinematics of the endoscope. The Depth evaluation metrics, mainly the absolute relative error (AbsRel), were adopted from~\cite{afsfm}. Before evaluation, depth maps were scaled following~\cite{sfm}. The Absolute trajectory error (ATE)~\cite{orb} was used for pose evaluation.

\section{Results}
We primarily compare Endo-FASt3r against SOTA reprojection-based SSL approaches for monocular depth and pose estimation ~\cite{sc-sfm,monodepth2,endo-sfm,yang,dares} in RAS.  To assess the generalisability of non-endoscopic foundation models DA V2 and Reloc3r, we also include their zero-shot performance. We use the reported metrics for all comparison methods in Table~\ref{restable}~\cite{dares,endodac}. To ensure fairness, we excluded SSL methods with extra modules beyond depth and pose in their reprojection, such as camera intrinsics estimation or image intrinsic decomposition, which add extra self-supervision signals~\cite{endodac,li2024image}. Even if included, Endo-FASt3r still excels significantly in pose estimation and is comparable or superior in depth estimation, proving its high performance without requiring additional parameters.
\newline
\newline
\newline
\newline

\begin{table}
\caption{Quantitive evaluation on the SCARED and Hamlyn datasets. ATE-T1/T2 denote ATE for trjactorries 1/2. Total/Train denote the total/trainable (millions) depth module parameters. Speed (ms) is reported for the depth module.}
\centering
\resizebox{1\textwidth}{!}{
\begin{tabular}{c|c|cccc|cc|cc|c}
\toprule 
\label{restable}
 & Method & AbsRel$\downarrow$ & SqRel$\downarrow$ & RMSE$\downarrow$ & $\delta\uparrow$ & ATE-T1 $\downarrow$ & ATE-T2 $\downarrow$ & Total & Train & Speed \\ \midrule 
\multicolumn{1}{c|}{\multirow{11}{*}{\rotatebox{90}{SCARED}}} & DeFeat-Net~\cite{defeat} & 0.077 & 0.792 & 6.688 & 0.941 &  0.1765 & 0.0995 & 14.8 & 14.8 & -  \\ 
& SC-SfMLearner~\cite{sc-sfm} & 0.068 & 0.645 & 5.988 & 0.957 & 0.0767 & 0.0509 & 14.8 &14.8 & -  \\ 
& Monodepth2~\cite{monodepth2} & 0.069 & 0.577 & 5.546 & 0.948 & 0.0769 & 0.0554 & 14.8 & 14.8 & - \\ 
& Endo-SfM~\cite{endo-sfm} & 0.062 & 0.606 & 5.726 & 0.957& 0.0759 & 0.0500 & 14.8 & 14.8 & -  \\
& AF-SfMLearner~\cite{afsfm} & 0.059 & 0.435 & 4.925 & 0.974 & 0.0757 & 0.0501 & 14.8 & 14.8 & 8.0 \\ 
& Yang et al.~\cite{yang} & 0.062 & 0.558 & 5.585 & 0.962 & 0.0723 & 0.0474 & 2.0 & 2.0 & - \\ 
& Zero-Shot DA V2 ~\cite{damv2} & 0.091 & 1.056 & 7.601 & 0.916 & - & - & - & - & - \\ 
% &  Fine-tuned DA V2~\cite{damv2} & 0.076 & 0.742 & 6.344 & 0.937 & - & - & 97.5 & 11.2 & 13.8\\ 
& Zero-Shot Reloc3r ~\cite{damv2} & - & - & - & - & 0.0938 & 0.0735 & - & - & - \\
% & EndoDAC~\cite{endodac} & 0.051 & 0.341 &  4.347 & 0.981& 0.0741 & 0.0512 & 99.0 & 1.6 & 17.7 \\ 
& DARES~\cite{dares} & 0.052 & 0.356 &  4.483 & 0.980 & 0.0752 & 0.0498 & 24.9 & 2.88 & 15.6 \\ \cline{2-11}
& \textbf{EndoFASt3r (Ours)} & \textbf{0.051} & \textbf{0.354} & \textbf{4.480} & \textbf{0.998} & \textbf{0.0702} & \textbf{0.0438} & 24.9 & 2.93 & 19.1 \\ \midrule
\multicolumn{1}{c|}{\multirow{3}{*}{\rotatebox{90}{Hamlyn}}} & Endo Depth \& Motion~\cite{endo-depth&motion} & 0.185 & 5.424 & 16.100 & 0.732& - & - & - & - & - \\ 
& AF-SfMLearner~\cite{afsfm} & 0.168 & ~\textbf{4.440} & 13.870 & 0.770 & - & -& 14.8 & 14.8 & 7.7 \\ \cline{2-11}
% & Fine-tuned DA V2~\cite{damv2} & 0.170 & \textbf{4.413} & 13.920 & 0.765 & - & - & 97.5 & - & 12.5 \\ \cline{2-11}
% & EndoDAC~\cite{endodac} & 0.138 & 2.796 & 11.491 & 0.813 & - & - & 99.0 & 1.6 & 15.7 \\
& \textbf{EndoFASt3r (Ours)} & \textbf{0.166} & 4.529 & \textbf{13.718} & \textbf{0.778} & - & - & 24.9 & 2.93 & 19.1 \\  \bottomrule 
\end{tabular}
} 
\label{tab: main}
\end{table}

\begin{figure}
\centering
\includegraphics[width=1\linewidth]{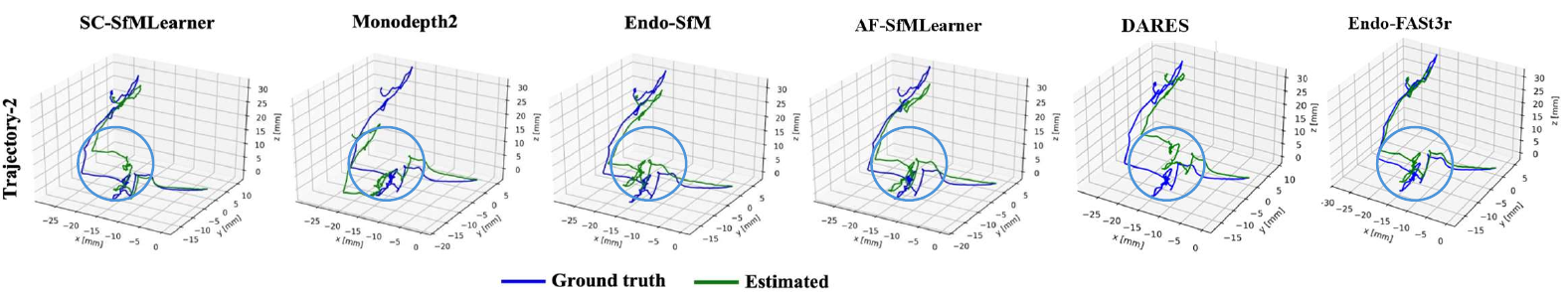}
\caption{Qualitative pose estimation comparison on the SCARED dataset}
\label{fig_pose}
\end{figure}

\begin{figure}
\centering
\includegraphics[width=1\linewidth]{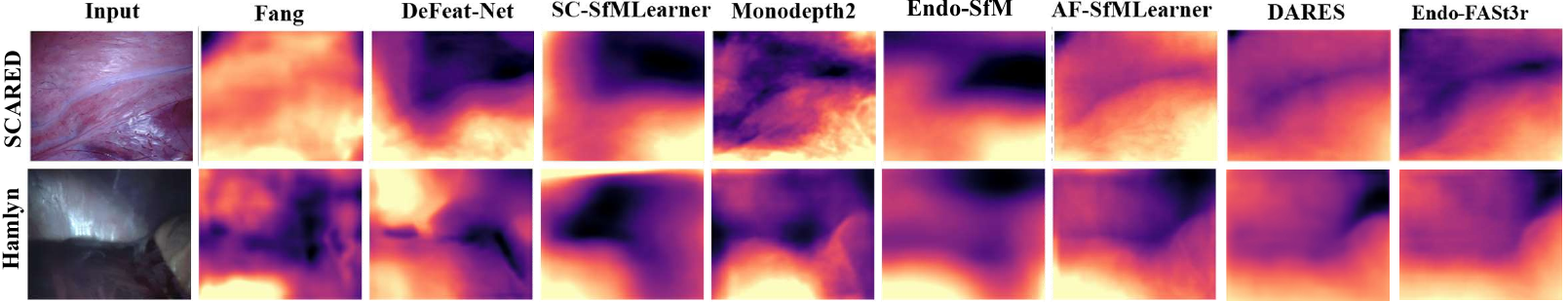}
\caption{Qualitative depth estimation comparison on the SCARED and Hamlyn datasets - Endo-Fast3r better captures edges demonstrating its robustness.}  
\label{fig_depth}
\end{figure}

\begin{figure}
\centering
\includegraphics[width=1\linewidth]{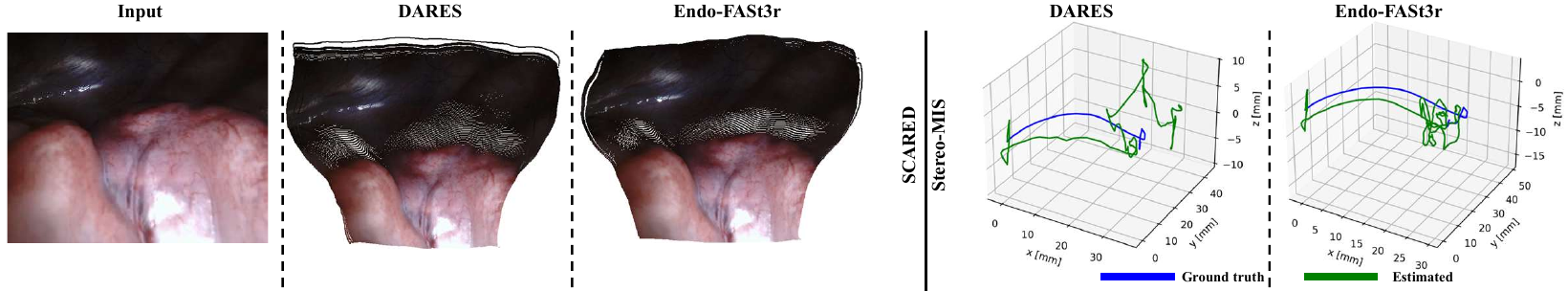}
\caption{Qualitative comparison with the second-best approach, DARES: SCARED 3D reconstruction (left) - StereoMIS pose estimation (right).}
\label{fig_3d}
\end{figure}

\noindent \textbf{Pose estimation:} Evaluation on the rigid SCARED dataset shows Endo-FASt3r outperforming all SOTA methods, improving by $6.64\%$ and $12.04\%$ - average $9.34\%$ - over the second-best approach, DARES (with a CNN-based pose module), for trajectories 1 and 2 respectively (Table~\ref{restable}). Fig.~\ref{fig_pose} presents this improvement on trajectory 2, where occlusion by surgical fluid is effectively addressed by Endo-FASt3r. To assess generalisability to a non-rigid case
with RAS-like tissue deformation and camera motion, we evaluated on the StereoMIS dataset, comparing against DARES, using its public code (Fig.~\ref{fig_3d}). Endo-FASt3r outperforms DARES by $7.13\%$ (Table~\ref{tab:stmis}). The pose module operates in real time, with a speed of $26ms$. Other works do not report this quantity. 
\newline
\noindent \textbf{Depth Estimation:} Evaluation on the SCARED dataset shows Endo-FASt3r outperforming all SOTA methods, surpassing the second-best approach, DARES by 2$\%$. While both use DA V2, DARES employs a LoRA-based adaptation, highlighting the advantages of DoMoRA over LoRA-based methods. Further testing on the Hamlyn dataset confirms Endo-FASt3r's effectiveness, with a 1.19$\%$ improvement over the next-best approach (Table~\ref{restable}). Depth estimation outputs from both datasets are shown in Fig.~\ref{fig_depth}. Notably, Endo-FASt3r only has $2.93$ million trainable parameters ($11.7\%$ of total parameters), and maintains a real-time inference speed of $19.1 ms$, slightly increased compared to SOTA methods. 

We also present 3D reconstructions on the SCARED dataset in Fig.~\ref{fig_3d}, which shows Endo-FASt3r better capturing darker areas with fewer artifacts.
\newline
\noindent \textbf{Ablation Studies:}
Table~\ref{tab:ablation} shows that DoRA and Reloc3rX improve pose estimation over the LoRA-based DARES (in the first row), while maintaining the same depth estimation accuracy, and convergence time despite Endo-FASt3r's increased number of parameters, thanks to DoRA's faster convergence properties. Using MoRA's full-rank matrices alone proves suboptimal. By integrating foundation models in both modules and leveraging both full-rank and low-rank updates, Endo-FASt3r enhances performance in both tasks while matching DARES in convergence time, despite having more parameters.

% \begin{table}
% \caption{Ablation study on the modules of Endo-FASt3r.}
% \centering
% \label{tab:ablation}
% \resizebox{0.70\textwidth}{!}{
% \begin{tabular}{c|c|c|c|cccc|cc}
% \hline 
% DoRA & MoRA & DoMoRA & Reloc3rX & AbsRel$\downarrow$ & SqRel$\downarrow$ & RMSE$\downarrow$ & $\delta\uparrow$ & ATE-T1$\downarrow$ & ATE-T2$\downarrow$\\ \hline 
% \xmark & \xmark & \xmark & \xmark & 0.052 & 0.366 & 4.522 & 0.979 & 0.0752 & 0.0498\\ 
% \checkmark & \xmark & \xmark & \checkmark & 0.052 & 0.487 & 4.875 & 0.968 & 0.0748 & 0.0505  \\ 
% \xmark & \checkmark & \xmark & \checkmark & 0.057 & 0.465 & 5.195 & 0.971& 0.0748 & 0.0505  \\  \hline 
% \xmark & \xmark & \checkmark & \checkmark & 0.051 &
% 0.354 & 4.480 & 0.998 & 0.0702 & 0.0438  \\ \hline 
% \end{tabular}} 
% \end{table}

\begin{table*}
\centering
\resizebox{1\textwidth}{!}{
\begin{floatrow}

% First Table: Ablation Study
\capbtabbox{
\resizebox{0.62\textwidth}{!}{
\begin{tabular}{c|c|c|c|cccc|cc}
\toprule
DoRA & MoRA & DoMoRA & Reloc3rX & AbsRel$\downarrow$ & SqRel$\downarrow$ & RMSE$\downarrow$ & $\delta\uparrow$ & ATE-T1$\downarrow$ & ATE-T2$\downarrow$\\ 
\midrule
\xmark & \xmark & \xmark & \xmark & 0.052 & 0.366 & 4.522 & 0.979 & 0.0752 & 0.0498 \\ 
\checkmark & \xmark & \xmark & \checkmark & 0.052 & 0.387 & 4.475 & 0.986 & 0.0735 & 0.0486 \\ 
\xmark & \checkmark & \xmark & \checkmark & 0.057 & 0.465 & 5.195 & 0.971 & 0.0748 & 0.0505 \\  
\midrule 
\xmark & \xmark & \checkmark & \checkmark & \textbf{0.051} & \textbf{0.354} & \textbf{4.480} & \textbf{0.998} & \textbf{0.0702} & \textbf{0.0438} \\ 
\bottomrule
\end{tabular}}
}{
\caption{Ablation studies on the SCARED dataset.}
\label{tab:ablation}
}

% Second Table: Pose Estimation
\capbtabbox{
\renewcommand{\arraystretch}{1.3}
\resizebox{0.28\textwidth}{!}{
\begin{tabular}{c|c|c}
\toprule
\multirow{3}{*}{\rotatebox{90}{StereoMIS}} & Method & ATE$\downarrow$ \\ 
\cline{2-3}
& DARES~\cite{dares} & 0.0715 \\ 
\cline{2-3}
& \textbf{EndoFASt3r (Ours)} & \textbf{0.0664} \\  
\bottomrule
\end{tabular}}
}{
\caption{StereoMIS results.}
\label{tab:stmis}
}
\end{floatrow}}
\end{table*}

\section{Conclusion}

We present Endo-FASt3r, a self-supervised monocular depth and pose estimation framework for RAS. 
We introduce Reloc3rX, the first development of a foundation model for pose estimation in endoscopic scenes. We also introduce DoMoRA, a novel adaptation technique that enables both low-rank and full-rank updates while benefitting from faster convergence. As the first SSL-based method that uses foundation models for both depth and pose estimation, Endo-FASt3r achieves a substantial improvement ranging from $7\%-10\%$ in the ATE metric for pose estimation and $2\%$ in the AbsRel metric in depth estimation over prior methods, as shown across three public datasets. Future work will explore the use of video foundation models~\cite{yang2024depth} to improve temporal stability.

\begin{credits}
	\subsubsection{\ackname} This work was supported in whole, or in part, by the the UKRI Centre for Doctoral Training (CDT) in AI-enabled Healthcare Systems [EP/S201612/1], the Department of Science, Innovation and Technology (DSIT), the Royal Academy of Engineering under the Chair in Emerging Technologies scheme, and the Engineering and Physical Sciences Research Council (EPSRC) under grants [EP/W00805X/1, EP/Z534754/1, UKRI145]. For the purpose of open access, the author has applied a CC BY public copyright licence to any author accepted manuscript version arising from this submission.
	
	\subsubsection{\discintname} The authors have no competing interests to declare that are relevant to the content of this article.
\end{credits}

\bibliographystyle{splncs04}
% \bibliography{ref}





% LNCS Homepage, \url{http://www.springer.com/lncs}, last accessed 2023/10/25
% \end{thebibliography}
\end{document}